**Méthodes pour la représentation informatisée de données lexicales/Methoden der Speicherung lexikalischer Daten**

Laurent Romary (Inria Saclay&Humboldt Universität zu Berlin)

Andreas Witt (Institut für Deutsche SpracheMannheim & Universität Heidelberg)

## Sommaire/Inhaltsverzeichnis

| | |
|---|---|
| Cet article à deux voix et en deux langues vise à offrir un état de l'art pédagogique sur la représentation de données lexicales informatisées. En traitant des mêmes thèmes en parallèle, sans pour autant que les deux textes soient des traductions l'un de l'autre, nous souhaitons ainsi pouvoir nous adresser à un large public de linguistes, lexicographes ou informaticien linguistes qui souhaitent connaître les bases de la modélisation et de la représentation de lexiques, qu'il s'agisse de dictionnaires ou de terminologies. Nous voyons ainsi ce chapitre comme un point de départ pour que nos lecteurs eux mêmes disposent des outils de représentations pour leur propre projet lexical. | In diesem Beitrag werden zwei Darstellungen zur Speicherung lexikalischer Daten in zwei verschiedenen Sprachen präsentiert. Die Texte beschreiben zwar in einer parallelenGliederung dieselben Themen, sind aber keine direkte Übersetzung voneinander. DiesesKapitel richtet sich an unterschiedliche Zielgruppen, neben Sprachwissenschaftler(inne)n und Lexikograph(inne)nrichtet er sich auch an Informatiker(innen) und Computerlinguist(inn)en, die mehr über die Grundlagen derModellierung undDarstellungvon digitalen Wörterbüchern lernen möchten. Wir betrachten dieses Kapitelals möglichen Ausgangspunktfür diejenigen, die lexikographische Projekte beginnen wollen und plädieren für eine gründliche Durchdringung der Problematik der Speicherung lexikalischer Daten. |

## 1. Introduction

L'arrivée des nouvelles technologies de l'information dans tous les domaines du processus de publication a influencé considérablement non seulement la constitution et l'élaboration de dictionnaires, mais a aussi changé en profondeur la vision attachée à la notion même d'entrée lexicale. Cette nouvelle perspective peut en particulier être imputée aux nouvelles possibilités de stockage, et ce faisant de représentation, qui se sont ainsi ouvertes pour les dictionnaires. La possibilité d'un stockage numérique systématique et surtout la mise en relation des sources de données les plus diverses ont changé fondamentalement le rapport initial au dictionnaire, dont la publication a jusqu'ici essentiellement reposé sur le livre papier.

On peut aussi observer que les premières représentations numériques associées à des dictionnaires ont surtout visé à offrir une visualisation des informations qui soit la plus proche possible de la version papier. Un exemple éminent d'un tel langage de description centré sur la notion de page est le format Postscript qui offrait déjà, autour des années 1980, la possibilité de sauvegarder simplement un dictionnaire sous une forme numérique qui reflétait la séquence des pages du livre. Ce format, développé par la société Adobe, et qui a précédé le toujours répandu PDF, peut encore être interprété, contrairement à de nombreux autres formats de cette époque, sur l'essentiel des ordinateurs actuels. Certes, une telle forme numérique offre déjà un certain nombre de fonctions de recherche qui n'étaient de toute façon pas disponibles sur les livres imprimés, mais l'exploitation maximale des contenus des dictionnaires informatisés ne peu se concevoir qu'en s'écartant résolument de la référence à la page imprimée (cf. Ide/Véronis 1995).

## 2. Quelques considérations techniques

Le présent chapitre ne s'adresse bien sûr pas à des spécialistes de l'informatique. Et pourtant, nous allons présenter, au fil des sections qui vont suivre, des éléments techniques qu'ils nous semblent important de maîtriser lorsque l'on manipule des données lexicales numériques. Afin de faciliter la compréhension de ces éléments techniques, nous proposonsdans cette section une vue d'ensemble du niveau où se positionne notre chapitre.

Si nous repartons de la notion de dictionnaire tel que nous le connaissons sous la forme d'un ouvrage imprimé, nous pouvons observer que les différentes entrées qu'il peut contenir possèdent, au moins pour les ouvrages modernes[1], une structure régulière découpant

---

[1]    Turcan, I.: «L'informatisation des premiers dictionnaires de langue française: les difficultés propres à la première édition du Dictionnaire de l'Académie française.», Les Dictionnaires de Langue française et

les différentes information en champs typographiquement bien identifiés (mot vedette, indications grammaticales, définition, exemples, étymologie, etc.). Nous pouvons alors essayer d'abstraire à partir de la structure observée et d'imaginer comment celle-ci pourrait être représentée sous forme numérique. Ce faisant, nous sommes en train d'identifier un *modèle* qu'il faudra ensuite traduire en termes strictement technique, sous la forme d'un *format* de représentation.

Tout d'abord, se pose le choix de la qualification des différents champs. Comme nous l'avons évoqué rapidement dans l'introduction, une possibilité est de nous appuyer sur un modèle qui traduit au plus prêt la présentation visuelle associée à chacun des champs. A chaque unité graphique, on associerait ainsi des propriétés de police de caractères, de taille de caractère, ou de propriété graphique (gras, italique, petites capitales) spécifiques. Cependant, une telle modélisation, bien que fidèle à la source papier et permettant effectivement, comme on le voit avec une représentation au format PDF, un enregistrement numérique, est de fait extrêmement limité du point de vue de l'exploitation que l'on peut faire de ces données. Il s'agit en effet non seulement de recouvrer ces données une fois enregistrées, mais aussi d'être en mesure d'effectuer des recherches (des requêtes) sue celles-ci, et au delà, de pouvoir définir des procédure de création, de maintenance et de diffusion dans des formats multiples (par exemple sur des tablettes graphiques ou sous la forme de pages web) à partir de ces mêmes données. Ce à quoi nous arrivons ici est qu'il nous faut définir un modèle traduisant la *structure logique* du dictionnaire à représenter, plutôt que sa *structure physique* telle qu'elle apparait sur la page papier.

L'étape suivante dans la réflexion liée à l'informatisation d'un dictionnaire est d'identifier une façon possible de traduire la structure logique de celui-ci sur un support numérique. La première chose qui vient à l'esprit, en particulier lorsque poussé par un financement que l'on a obtenu, et de s'adresser à un informaticien pour qu'il conçoive une „base de données" qui va pouvoir mettre en œuvre le projet. Dois s'engager alors un dialogue entre le lexicographe et l'informaticien pour que celui-ci comprennent les contraintes portant sur les données et que le lexicographe quand à lui puisse être convaincu que ses besoins sont bien pris en compte. Quelque soit la qualité du produit final, il se traduit toujours par une certaine frustration vis-à-vis de ce que l'on a ainsi obtenu. En effet, la base de donnée reste un objet opaque dont on sent bien que sans l'informaticien, on ne saura jamais en faire évoluer la structure. Par ailleurs, on a l'impression que le logiciel et les données avec, sont difficilement comparables avec le travail qu'ont pu faire d'autres équipes lexicographiques similaires, sic e n'est en comparant vaguement les interfaces en ligne disponibles.

Dans ce cadre, le présent papier adopte une approche à mi-chemin entre la connaissance lexicographique abstraite et le développement lourd d'une base de donnée, en affirmant qu'il est nécessaire pour le lexicographe d'avoir les moyens de définir minimalement ses besoins dans un langage formel qui puisse être à la base de l'environnement informatique futur qui l'implémentera, et ce en cohérence avec les normes internationales qui justement régissent la façon de modéliser et de représenter des données lexicales informatisées. La base de cette approche est l'utilisation de la norme XML[2] définie par le consortium W3C[3] et qui offre un cadre descriptif de représentation de données textuelles sous la forme de balises qui forment tune structure hiérarchique de parenthésage. Par exemple, au niveau leplus simple, on pourra marquer que la prononciation du mot 'chat' dans une entrée de dictionnaire ainsi: <pron>ʃa</pron>, créant ainsi ce qu'on appelle un *élément* XML, l'unité de base de telles représentation. Sans faire ici une présentation exhaustive de ce langage — on verra qu'il est relativement facile de s'y retrouver en observant directement les exemples fournis — nous pouvons signaler quatre aspects importants qui impactent sur le rôle que peut jouer XML dans un processus éditorial lexicographique:

– Tout d'abord des données exprimées en format XML sont directement lisibles par tout éditeur de texte mais peuvent être aussi manipulées de façon extrêmement précise à l'aide d'éditeur XML spécialisés ;

– La structure d'un document XML peut être contrôlée à l'aide d'un langage formel, appelé *schéma*, dont la spécification va être au centre du travail de modélisation du lexicographe ;

– La représentation de données en XML peut aussi être vue comme une interface avec une base de donnée complexe qui va lire et produire de telle donnée, et offrir ainsi un environnement puissant et efficace de gestion des informations ;

– Enfin, les données XML peuvent être facilement transformées et mises en ligne à l'aide des standards idoines (XSLT et CSS notamment) définis par le W3C. Ces langages permettent une diffusion lisible des données lexicographiques en complément de la diffusion des données sources XML proprement dites.

Dans la suite de ce chapitre, nous adopterons XML comme base de représentation finale des données en analysant les classes de modèles lexicaux sous-jacents qui vont nous permettre de caractériser tel ou tel format particulier.

---

[2] Extensible Markup Language: http://www.w3.org/TR/2006/REC-xml11-20060816/
[3] World Wide Web Consortium: l'instance de normalisation des formats du web

## 3. Modèles lexicaux

### 3.1 Deux modèles complémentaires

La représentation de données lexicales, que ce soit sous une forme traditionnelle imprimée ou de façon numérique, correspond toujours à l'application d'un modèle sur le matériau linguistique.Ce modèle détermine la façon dont on conçoit la notion de lexème par exemple, la relation d'un lexème à son ou ses sens, ainsi que le niveau de granularité auquel on souhaite décrire ces différents éléments. Dans cette section, nous présentons deux classes de modèles génériques qui couvrent l'essentiel des pratiques de représentation lexicale. Ces classes de modèles, on parle parfois de méta-modèle, sont en général affinés par des choix éditoriaux plus précis en fonction des objectifs visés ou de la disponibilité suffisante d'informations.

Les deux classes de modèles, nommées respectivement *sémasiologique* et *onomasiologique*, correspondent au deux grandes façons d'articuler forme et sens dans une représentation lexicale.

Les modèles sémasiologiques correspondent à des représentations où l'élément de départ est le lexème, pour lequel on indique dans un deuxième temps le ou les sens (*sema-*). Ils visent en général à décrire un répertoire linguistique assez large tels qu'on peut le rencontrer dans les dictionnaires traditionnels. La finesse de représentation des sens, éventuellement sous la forme d'une hiérarchie de sous-sens dépend de l'approche lexicographique envisagée.

Les modèles onomasiologiques au contraire supposent que l'on dispose d'un répertoire de sens — on parlera plutôt de *concepts* — pour répertorier ensuite, éventuellement dans plusieurs langues, les différentes formes linguistiques (parfois qualifiés de *termes*) qui peuvent faire référence à ce concept. Ces modèles reposent donc sur une hypothèse très particulière que l'on rencontre en générale dans les langues de spécialité, ce qui explique leur utilisation par les traducteurs ou les rédacteurs techniques.

De part leurs approches clairement différentes, et souvent complémentaires pour de nombreuses applications concrètes, les modèles onomasiologiques et sémasiologiques sont plus ou moins à même de prendre en compte des phénomènes lexicaux de bases telles que la polysémie ou la synonymie. Dans les sections suivantes, nous présentons plus précisément comment cette opposition se traduit dans les faits.

## 3.2 Polysémie

La complémentarité ou l'opposition entre modèles sémasiologique et onomasiologique se traduit particulièrement bien lorsque l'on observe comment l'un et l'autre se positionnent vis-à-vis de la représentation de la polysémie.

Par définition, les modèles sémasiologiques intègrent la polysémie dans leur organisation, puisque pour chaque entrée lexicale, on répertorie les différents sens qui y sont associés. Ce découpage en sens, qui peut correspondre à différentes pratiques éditoriales (granularité, récursivité éventuelle des sens), traduit explicitement le niveau de polysémie de l'entrée lexicale correspondante.

Dans le cas des modèles onomasiologiques, aucun mécanisme n'existe pour représenter la polysémie, puisqu'un terme est considéré a priori comme faisant référence à un et un seul concept dans un domaine de spécialité particulier. La polysémie n'apparait que de façon "accidentelle" quand une base terminologique multi-domaine contient plusieurs entrées auxquelles sont rattachées le même terme.

## 3.3 Synonymie

La représentation de la synonymie dans les deux modèles lexicaux repose sur un phénomène dual de celui observé dans le cas de la polysémie.

Dans le modèle onomasiologique la polysémie fait partie de la construction même d'une entrée conceptuelle, puisque tous les termes intégrés à une entréepour une même langue sont considérés comme synonymes puisqu'ils expriment ce même concept. Dans le cas multilingue, les termes qui apparaissent dans le même concept, mais dans des langues différentes sont considérés comme des équivalents de traduction.

L'autonomie du terme dans le modèle sémasiologique complet, permet cependant d'affiner cette relation synonymique, soit par la qualification du terme lui-même (terme préféré, registre, conditions d'emploi, etc.), soit par l'ajout de relations entre termes (équivalence de traduction explicite, relation de dérivation, abréviation, etc.).

La notion de polysémie est quant à elle a priori absente de la démarche sémasiologique, puisque l'on y décrit les sens possibles d'un mot ou d'une expression indépendamment des sens attachés aux autres entrées dans le lexique. Cependant, les dictionnaires ont classiquement représenté des liens d'équivalence entre entrées lexicales, soit globalement, soit au niveau des sens, par le biais de renvois explicites (*synonyme de*, *voir aussi*, etc.). Dans une représentation informatisée, ces renvois se traduisent par des pointeurs plus ou moins typés, et dans les modèles plus élaborés (tels que Papillon [Sérasset et al. 2000], ou le module de traduction de la norme LMF) par la réification, c'est à dire l'explicitation, de liens d'équivalence.

Un cas particulier que l'on peut mentionner ici est celui de la représentation de variantes d'une entrée lexicale donnée, qu'il s'agisse de variantes orthographiques (cheick/cheik) ou de la représentation d'un acronyme avec sa forme intégrale (ONU/Organisation des Nations Unies). Le choix éditorial qui se pose ici dans le cas d'un modèle sémasiologique est soit de créer une entrée pour chacune des variantes et de gérer des renvois vers celle de ces entrées où l'essentiel des descriptions lexicales est réuni, ou bien d'intégrer dans une seule et même entrée l'information qu'il existe une alternance des formes de surface. Dans le cas onomasiologique, comme on l'aura compris, il suffira d'intégrer les différentes formes comme équivalentes pour une entrée donnée. On pourra, comme on le verra bien sûr marquer la nature lexicale (par exemple qu'il s'agit d'un acronyme) en association avec un terme particulier.

## 3.4 Aspects multilingues

La représentation de données lexicales multilingues dans les modèles sémasiologiques et onomasiologiques reflète les observations faites au sujet de la synonymie.

Dans la cas sémasiologique, on ne peut traduire des équivalences multilingues qu'à l'aide de renvois explicites d'un lexique vers un autre, de sorte que pour un mot donné dans une langue source, ou l'un de ses sens, on indique une analogie ou une équivalence vers une entrée lexicale, ou l'un de ces sens, dans un lexique de la langue cible. Ce mécanisme rend d'une part difficile la maintenance de relations cohérentes dans les deux sens linguistiques, et par ailleurs ne permet quasiment pas d'étendre le nombre de langues du fait de la complexité du réseau de lien que cela entrainerait.

Tout au contraire, le modèle onomasiologique est parfaitement adapté à la représentation de données multilingues. Pour un concept donné, le modèle permet non seulement de répertorier l'ensemble des termes permettant de l'exprimer dans une langue donnée, mais tout aussi bien d'effectuer cette opération pour un nombre illimité de langues, dès lors que le concept puisse y être exprimé. Dans le cas le plus simple, où l'on a des équivalences bilingues, on obtient ainsi une structure telle qu'illustrée ci-dessous:

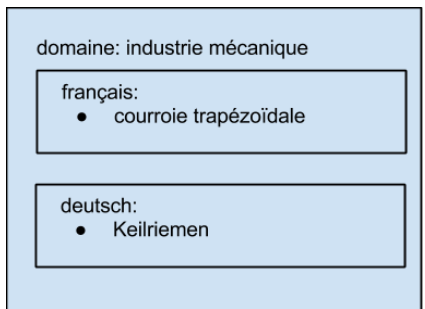

**Figure 1:** Entrée multilingue dans une représentation onomasiologique

## 3.5 Couverture lexicale et domaines d'applications

Pour conclure sur la description des propriétés principales des deux modèles sémasiologiques et onomasiologiques, on peut observer qu'ils correspondent de fait à deux classes d'applications très différentes. Le fait qu'un lexique sémasiologique permette facilement de répertorier un ensemble large de lexèmes est particulièrement adapté aux applications où la couverture lexicale est un aspect essentiel. Typiquement, un dictionnaire de référence, mais aussi un lexique simple de formes fléchies pour le traitement des langues, vont reposer sur cette classe de modèles. Inversement, l'organisation conceptuelle, ainsi que la capacité de représenter facilement des données multilingues, vont favoriser l'usage des modèles onomasiologiques pour les applications liées à des domaines techniques: rédaction technique, traduction ou indexation de textes. On remarquera que dans les faits, les couvertures lexicalesque l'on observe pour les lexiques reposant sur l'un ou l'autre modèle sont très différentes. Les lexiques sémasiologiques, à l'exemple des dictionnaires traditionnels, vont couvrir l'essentiel des formes simples d'une langue, et ce pour toutes les parties du discours, alors que les lexiques onomasiologiques, sous la forme principalement de terminologies multilingues, vont intégrer un grand nombre de formes complexes, plutôt nominales (occasionnellement verbales ou adjectivales), et organisées par domaine de spécialité.

## 4. Des modèles aux formats

Dès lors que l'une ou l'autre des approches exposées ci-dessus a été choisie, il convient alors de déterminer quelles sont les possibilités de traduire exactement les données lexicales sous la forme de représentations numériques. Il faut pour cela développer un modèle formel qui puisse refléter exactement les différentes unités d'information du modèle théorique. Une fois que ce travail de modélisation est effectué, il est relativement simple, comme nous le verrons par la suite, d'instancier le modèle dans une représentation concrète, notamment en format XML. Dans ce cadre, et indépendamment de l'approche (sémasiologique ou onomasiologique) choisie, nous allons présenter une méthodologie commune qui va guider le processus de modélisation et de détermination d'un format numérique. Cette méthodologie repose sur trois éléments principaux:

– La présentation d'un *méta-modèle* qui va décrire les grandes composantes communes à toutes les représentations de données pour l'une ou l'autre des classes lexicales (onomasiologique ou sémasiologique) que nous avons décrites précédemment ;

– La fourniture d'une base de descripteurs, que nous appellerons *catégories de donnée*, qui vont fournir des unités d'information associées aux différents composants d'un méta-modèle donné. Par exemple, la catégorie de donnée *genre grammatical* pourra être associé à la description d'un terme dans une entrée onomasiologique;

– Enfin, la détermination d'un cadre pour la traduction des modèles obtenus dans le langage XML, ce que nous appellerons *sérialisation*.

Pour chacun des ces éléments, nous nous appuierons sur les travaux menés depuis plusieurs années dans le cadre de la normalisation internationale. Cette référence aux normes est particulièrement importante dans la perspective de garantir à la fois une réutilisation transparente des données ainsi représentées, mais aussi une pérennisation de celles-ci, de sorte qu'elle restent lisibles et exploitables dans le temps, indépendamment de l'évolution des matériels et logiciels informatiques. Nous nous appuierons plus particulièrement sur deux cadres de normalisation:

– Le comité technique 37 de l'ISO, où les méta-modèles TMF[4] (onomasiologique) et LMF[5] (sémasiologique) ont été définis, ainsi que le cadre de définition de catégories de données normalisés ;

– Lc consortium TEI (Text Encoding Initiative), qui depuis 1987 définit et gère un ensemble de formats XML pour la représentation de données textuelles.

La figure ci dessous synthétise les éléments qui vont être décrits plus précisément dans la suite de ce chapitre.

---

[4]  Terminological Markup Framework
[5]  Lexical Markup Framework

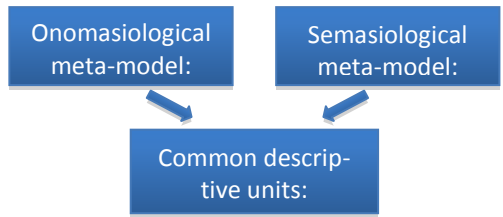

**Figure 2:** Organisation générale des standards pour la représentation de données lexicales

## 5. Standards et formats pour les modèles onomasiologiques

### 5.1 Principes de base

La représentation de données terminologiques conformes à la démarche onomasiologique repose sur deux grands principes fondamentaux qui sont à la base des normes et formats que nous allons décrire dans cette section (cf. Wright/Budin 1997):

– selon le principe de l'orientation conceptuelle (*concept orientation*), l'ensemble de l'information terminologique relative à un concept donné est regroupée dans une même représentation; sont en particulier concernés l'ensemble des termes pouvant faire référence au concept,dans toutes les langues considérées, ainsi que toute information descriptive ou administrative qu'il est nécessaire d'y associer;

– le principe de l'autonomie du terme (*term autonomy*) stipule que chaque terme associé à un concept doit être géré comme une unité descriptive autonome, sans qu'aucun terme n'ait une position particulière dans le modèle.

### 5.2 Le méta-modèle TMF

La norme ISO 16642 (TMF – Terminological Markup Framework) est le document de référence pour la description de modèles de représentation de données lexicales onomasiologiques, principalement dans le domaine des terminologies multilingue. Comme illustré ci dessous sur la base de la base de l'exemple déjà utilisé ("Keilriemen"), le modèle organise une entrée onomasiologique sous la forme d'une représentation à trois niveaux:

– le niveau concept (*Terminological Entry* – que nous abbrévierons TE) représente une entrée élémentaire dans une base terminologique. C'est là que vont être représentées les informations descriptives de base relatives au domaine de spécialité, à la signification générale du concept (définition), ainsi que d'éventuelles relations (hyperonimiques ou méronimiques) avec d'autres concepts;

- le niveau langue (*Language Section* –LS) regroupe l'ensemble des descriptions relatives à une "langue" donnée, en fonction du choix éditorial de la base terminologique (ex. différentiation des régionalismes, représentation des variations dialectales, etc.).
- le niveau terme (*Term Section* – TS) regroupe toutes les informations descriptives propres à un terme particulier dans une langue donnée, et ce en conformité avec le principe d'autonomie du terme énoncé ci-dessus.

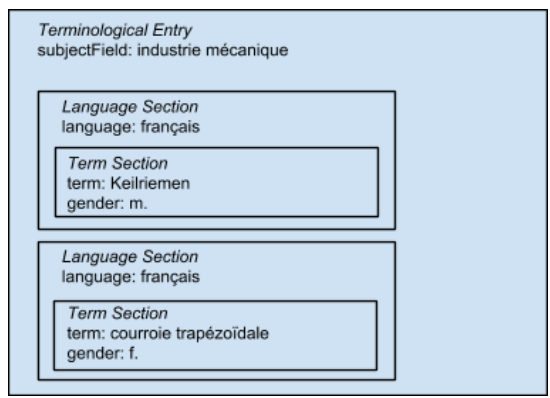

**Figure 3:** Illustration du méta-modèle TMF (ISO 16642)

La majorité des catégories de données susceptibles d'apparaître dans une entrée terminologique est décrite dans l'ancienne norme ISO 12620:1999. Cependant, dans la plupart des applications, seul un sous-ensemble restreint est véritablement utilisé. Nous décrivons ici ces principales catégories en les reliant aux composants du méta-modèle TMF.

Pour disposer d'un modèle terminologique valide, seules deux catégories de données sont obligatoires, car elles déterminent la création des composants correspondants dans le méta-modèle TMF:
- "term": qui contient la forme de référence (lemme) du terme au sein du composant *Term Section*;
- "language": qui détermine la langue objet d'un composant *Language Section*.

Un modèle reposant sur ces deux seules catégories permet typiquement de représenter des listes d'équivalents multilingues.

Pour décrire des entrées terminologiques plus élaborées, différentes catégories de données issues de la norme ISO 12620:1999 peuvent être utilisées, mais trois jouent un rôle particulier et sont en général intégrés à tout modèle terminologique:
- "subjectField" permet de décrire le domaine de spécialité de l'entrée terminologique (au niveau du composant *Terminologic Entry*) et donc de structurer une base terminologique en champs thématiques;

- "definition" permet de donner une définition simple (au sens de la norme ISO 704) pour le concept sous-jacent à l'entrée terminologique. Cette catégorie est généralement utilisée au niveau de *Terminological Entry*, mais il arrive, notamment dans le cas de bases multilingues qui doivent pouvoir être utilisées quelque soit l'origine linguistique du lecteur, qu'on l'intègre au niveau *Language Section*;
- enfin, "partOfSpeech" (avec éventuellement "gender") accompagne naturellement le terme au niveau du composant *Term Section* pour indiquer la catégorie grammaticale du terme.

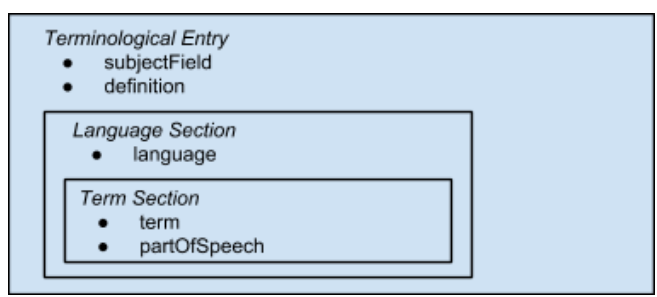

**Figure 4:** Ajout de catégories de donnée au modèle TMF

A partir d'une telle base, des catégories de données additionnelles peuvent être utilisées pour affiner ou qualifier les représentations aux trois niveaux considérés. Il est ainsi possible d'ajouter des informations administratives concernant les sources utilisées, les responsabilités ou les dates de création et de modification des entrées à tous les niveaux du modèle. On peut de même qualifier les termes en fonction de leur registre, de leur degré d'autorité ou de leur statut linguistique (e.g. abréviation). Enfin, le modèle permet l'ajout d'exemples attestés pour un terme donné.

## 5.3 Sérialisation de TMF à l'aide du format TBX

La norme ISO 16642 (TMF) est en fait le fruit de longs développements normatifs qui ont débuté dans les années 80 avec la publicaiton de la norme ISO 6156:1987 (Mater) et permettant d'aboutir à la publication en 2008 de la norme ISO 30042 (TBX). TBX se présente de fait comme une implémentation de la norme TMF en offrant une sérialisation de référence extrèmement modulaire, permettant à toute application de définir le format dont elle peut avoir besoin tout en restant compatible avec la norme principale.

La partie centrale du méta-modèle de TMF, à savoir les composants *Terminological Entry*, *Language Section* et *Term Section* sont implémentés dans TBX à l'aide des éléments `<termEntry>`, `<langSec>`et `<tig>` respectivement. Si on y ajoute l'élément `<term>`pour implémenter la catégorie de donnée correspondante, on obtient une structure minimale qui est illustrée ci-dessous:

```xml
<termEntry xml:id="c5" xmlns="http://www.tbx.org">
<langSet xml:lang="en">
<tig>
<term>e-mail</term>
</tig>
</langSet>
<langSet xml:lang="fr">
<tig>
<term>courriel</term>
</tig>
</langSet>
</termEntry>
```

Les autres catégories de données sont implémentées à l'aide d'éléments génériques couplés à un attribut `@type` qui en affine la sémantique. Ces éléments sont au nombre de trois:

- `<descrip>`: permettant d'apporter des informations descriptives liées à l'entrée terminologique, par exemple le domaine ou une définition ;
- `<admin>` pour l'ajout d'informations administratives sliées aux différents niveaux d'une entrée terminologique, par exemple la base de donnée d'origine, ou l'identifiant associé au niveau de représentation ;
- `<termNote>`, qui permet de spécifiquement qualifier le term au sein d'un block `<tig>`.

Dans l'exemple suivant[6], nous voyons comment ces différents éléments sont utilisés en combinaison avec des catégories de données précises.

```xml
<termEntry>
<descripGrp>
        <descrip type="originatingDatabaseName">Vocabulaire multidisciplinaire PAS-
        CAL</descrip>
<descrip type="subjectField" xml:lang="fr">Biomédical</descrip>
<admin type="conceptIdentifier">BV.122497</admin>
<admin type="conceptOrigin">INIST</admin>
</descripGrp>
<langSet xml:lang="fr">
<tig>
<term>Acide 1,2-dithiolane-3-valérique</tei:term>
<admin type="termIdentifier">BV.122497.1</admin>
<termNote type="administrativeStatus">preferredTerm</termNote>
</tig>
<tig>
<term>«1,2»-Dithiolane-«3»-valérique acide</tei:term>
<admin type="termIdentifier">BV.122497.2</admin>
<termNote type="administrativeStatus">deprecatedTerm</termNote>
</tig>
<tig>
<term>Acide α-lipoïque</tei:term>
<admin type="termIdentifier">BV.122497.3</admin>
</tig>
<tig>
<term>Acide thioctique</tei:term>
<admin type="termIdentifier">BV.122497.4</admin>
</tig>
</langSet>
<langSet xml:lang="en">
<tig>
```

---

[6] Issu des travaux menés au sein du projet ANR TermiTH en collaboration avec l'INIST-CNRS, que nous remercions ici.

```
<term>Thioctic acid</tei:term>
<admin type="termIdentifier">BV.122497.5</admin>
<termNote type="administrativeStatus">preferredTerm</termNote>
</tig>
</langSet>
</termEntry>
```

## 6. Standards et formats pour les modèles sémasiologiques

### 6.1 Introduction

Comme nous l'avons vu, les modèles semasiologiques correspondent à une très grande variété d'applications et de ce fait couvrent une large gamme de modèles possibles. Dans sa version la plus simple, une entrée sémasiologique peut se réduire à la description d'une ou plusieurs formes, comme c'est le cas dans les lexiques flexionels (*full-form lexica*). A l'opposé, les lexiques destinés au traitement automatique des langues peuvent contenir des descriptions complexes telles que les constructions syntaxiques pouvant être associées à une entrée. Dans cette section, nous mettrons l'accent sur les modèles permettant de représenter ce que l'on qualifierait plutôt de "dictionnaire", qu'il s'agisse de dictionnaires imprimables classiques ou en ligne.

### 6.2 Présentation générale (et rapide) de LMF

La norme LMF est l'aboutissement d'un travail de concertation entre plusieurs groupes internationaux pour définir un modèle générique de représentation de données lexicales sémasiologiques. Il s'appuie notamment sur les travaux effectués dans le cadre de différents projets européens tels que Multext[7], Parole et Simple. LMF vise en fait à être une plate-forme de spécification de données lexicale permettant de représenter tout type de forme lexicale (sémasiologique) particulière, qu'il s'agisse de dictionnaires à usage humain (MRD – Machine Readable Dicitonnary), ou de lexiques destinés au traitement automatique des langues: lexiques flexionnels, lexiques syntaxiques, etc.

La norme LMF a une approche modulaire en considérant que tout lexique peut être modélisé à partir d'un noyau commun, complété d'une ou plusieurs extensions spécifiques à l'application concernée. Le noyau commun, illustré dans la figure ci-dessous, traduit la perspective sémasiologique en considérant qu'une entrée lexicale (*Lexical Entry*) est en premier lieu caractérisée par une ou plusieurs formes (*Form*), par exemple la forme correspondant au lemme et celles associées aux formes fléchies ou variantes orthographiques. Ces formes sont elles-mêmes réalisées par le biais d'une ou plusieurs representations (*Form Representation*)

---

de façon à grouper les représentation phoniques, orthographiques ou translitérées dans une même structure.

Une fois l'entrée lexicale caractérisée par le composant *Form*, le noyau du modèle LMF propose une simple description du contenu sous la forme d'un composant *Sense*, qui peut être répété et décomposé récursivement en sous-senses.

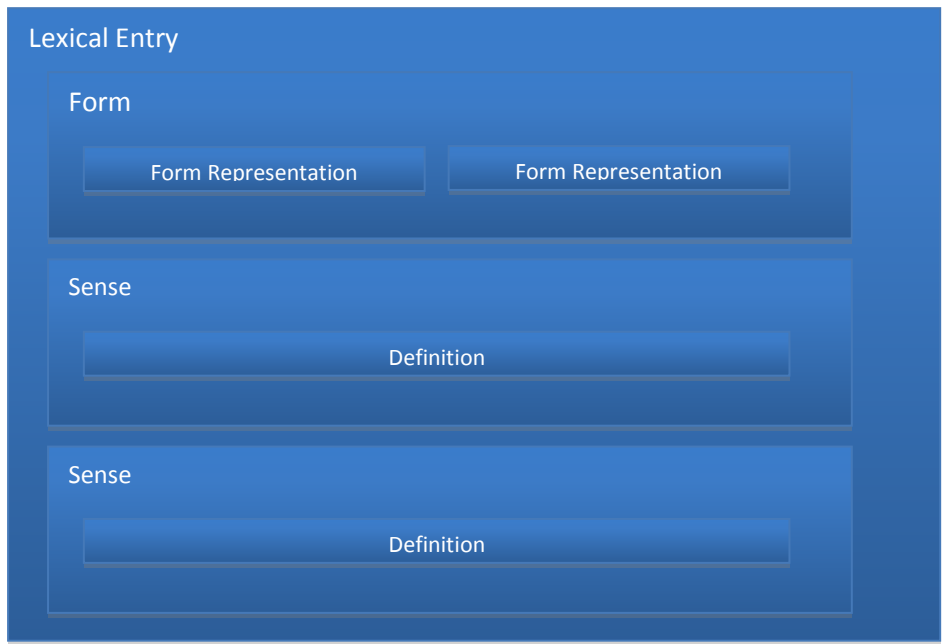

**Figure 5:** Méta-modèle noyau de la norme LMF

En complément du modèle noyau, LMF propose une extension spécifique pour les diction­naires informatisés. Cette extension prévoit, en complément du composant *Definition*, la re­présentation d'un ou plusieurs exemples d'usage à l'aide du composant *Context*. Ces exemples peuvent apparaître sous différentes formes (systèmes d'écriture, translitération ou prononciation). Bien que la définition de ce composant dans la norme LMF prévoit que l'entrée lexicale puisse apparaître sous une forme fléchie, aucun mécanisme n'est fourni pour ce faire.

Le deuxième composant introduit par l'extension MRD est *Subject Field* qui permet de donner une indication simple ou complexe (le composant peut être itéré et est lui-même récursif) de domaine pour le sens correspondant. Enfin, dans le cas des dictionnaires multi­lingues, l'extension introduit aussi un composant destiné à la fourniture de traductions dans une autre langue: *Equivalent*.

### 6.3 Sérialisation TEI et mécanismes d'extension

Le chapitre *Dictionnaires* ("Dictionaries") des directives de la TEI est un candidat naturel pour la représentation de dictionnaires conformes au méta-modèle LMF. Comme nous le signalons dans (Romary, à paraître), il y a deux raisons principales qui rendent naturelle la sérialisation en TEI :

– le modèle du chapitre *Dictionnaires* de la TEI a été conçu initialement comme une organisation générique d'entrées lexicales sémasiologiques et a de ce fait influencé fortement les premières réflexions autour de la norme LMF ;

– bien que couvrant un spectre beaucoup plus large de modèles lexicaux les mécanismes de customisation de la TEI, permettent d'identifier un sous-ensemble qui correspond au mieux au méta-modèle LMF ;

Par ailleurs, la TEI offrent des possibilités bien plus fines que ce qui est prévu dans LMF à la fois pour associer des méta-données à une base lexicale et pour annoter finement les contenus textuels (tels que les définitions ou les exemples). Nous reviendrons sur ce dernier point à la fin de la présente section.

Une entrée de dictionnaire se décrit dans les directives de la TEI à l'aide de l'élément `<entry>`, qui peut apparaître à tout niveau de division d'un document TEI, éventuellement en alternance avec d'autres bloc textuels (paragraphes, listes, descriptions bibliographiques etc.). L'organisation du méta-modèle LMF en *Form* et *Sense* est directement implémentée à l'aide des deux éléments `<form>` et `<sense>`.

L'élément `<form>` peut contenir différent sous-éléments qui vont pouvoir sérialiser le composant *Form Representation* en fonction de ce qu'il doit représenter : `<orth>` pour les formes orthographiques, `<phon>` pour la prononciation, `<hyph>` pour les indications de rupture typographique, `<stress>` pour l'indication de schémas d'accentuation ou `<syll>` pour la syllabification. La différence entre le lemme et les autres formes d'une entrée s'implémente en TEI par un attribut `@type` sur l'élément `<form>` (typiquement, avec les valeurs 'lemma' et 'inflected' pour représenter un lemme et ses formes fléchies).

Il possède par ailleurs deux caractéristiques qui, bien que non prévues dans le méta-modèle LMF, complémente bien celui-ci :

– Les informations grammaticales associées à une forme peuvent être regroupées au sein d'un même élément `<gramGrp>`, ce qui facilite leur gestion et leur interrogation ;

– L'élément `<form>` est lui-même récursif, ce qui offre un moyen simple de traiter les expressions lexicales complexes.

De la même façon, l'élément `<sense>` de la TEI a une structure très proche du composant *Sense* de la norme LMF. Cet élément est récursifet implémente l'ensemble des composants prévus dans le modèle noyau de LMF, ainsi que dans l'extension MRD:

– L'élément `<def>`implémente le composant *Definition* de LMF, mais peut contenir des annotation de surface bien plus fines;

– La représentation d'exemples dans la TEI, correspondant au composant *Context* de LMF, repose sur l'élément générique `<cit>`, qui permet de décrire tout type de réalisation linguistique, qu'il s'agisse d'exemple d'usage ou de traduction. Cet élément enrichit nettement la norme LMF, puisqu'il permet d'associer à tout exemple des informations particulières relatives à sa source bibliographique, les conditions d'usage spécíque ou en associant cet exemple lui-même à une traduction;

– Enfin, le composant *Subject Field* est implémenté à l'aide de l'élément `<usg>` (*usage*) qui en fait le généralise lui aussi puisque `<usg>` permet, par le biais d'une surspécification à l'aide d'un attribut `@type`, de couvrir tout indication d'usage diastratique (correspondant au composant *Subject Field*: `<usg type='dom'>`),diachronique (`<usg type='time'>`), diatopique (`<usg type='geo'>`) ou diaphasique (valeurs 'register' ou 'style' de `@type`).

Ces différents mécanismes sont illustrés ci-dessous à l'aide d'un exemple complet d'entrée lexicale, inspiré des directives de la TEI, mais rendu conforme pour recouvrir exactement le méta-modèle LMF.

```
<entry>
<form>
<orth>poussin</orth>
<pron>pusë</pron>
</form>
<gramGrp>
<pos>n.</pos>
<gen>m.</gen>
</gramGrp>
<sense n="1">
<def>Jeune poulet, nouvellement sorti de l'oeuf, encore couvert de duvet.</def>
<cit type="example">
<quote>La poule et ses poussins.</quote>
</cit>
</sense>
<sense n="2">
<usg type="dom">Zool.</usg>
<def>Jeune oiseau (par rapport aux adultes, aux parents)</def>
```

```
</sense>
<sense n="3">
<usg type="dom">êtres humains</usg>
<sense n="3.1">
<usg type="register">Fam.</usg>
<def>Terme d'affection (enfant)</def>
</sense>
<sense n="3.2">
<usg type="dom">Sports</usg>
<def n="2">Catégorie d'âge (9 ans) qui précède celle des benjamins</def>
</sense>
<sense>
<def n="3">Elève de première année dans certaines écoles (Air, Aéronauti-
que)</def>
</sense>
</sense>
</entry>
```

## 7. Et maintenant...

Ce chapitre doit être vu comme une base conceptuelle et méthodologique pour quiconque veut représenter des données lexicales informatisées. Les différentes normes présentées ici doivent servir de base à tout nouveau projet lexical pour garantir, d'une part, que les formats adoptés reposent sur une organisation cohérente des unités d'information et, d'autre part, que les données obtenues soient comparables avec d'autres projets similaires. Il est des domaines ou trop de créativité peut devenir contre-productive et nous encourageons nos lecteurs à ne pas considérer que le travail d'assimilation des normes existantes puisse être superflu.

## 1. Einleitung

Wörterbücher sind Bücher. Dieser Satz klingt tautologisch und war es lange auch, aber heute ist er nicht mehr notwendigerweise richtig.[8] Der Einzug neuer Informationstechnologien in alle Bereiche des Publikationsprozesses beeinflusste nicht nur die Erstellung und Bearbeitung von Wörterbüchern erheblich, sondern führte auch zu der neuen Produktgattung des digitalen Wörterbuchs.

Grundsätzlich bieten bereits klassische Wörterbücher, wenn sie digital gespeichert werden, erweiterte Nutzungsmöglichkeiten, insbesondere zur Suche. Die hierbei eingesetzten Datenformate orientieren sich am Erscheinungsbild der Seiten des bedruckten Papiers. Ein prominentes Beispiel für ein solches Format ist die Seitenbeschreibungssprache PostScript (Adobe 1987), die schon seit den 1980er Jahren die Möglichkeit bot, ein Wörterbuch in einem digitalen Format zu speichern, das eine Sequenz der digitalen Repräsentation der Seiten des Buches bildete. Dieses von der Firma Adobe entwickelte Format und dessen – inzwischen sehr weit verbreiteter– Nachfolger,das Format PDF (ISO 32000-1:2008), können– imGegensatz zu einer Vielzahl anderer Formate dieses Alters – nachwie vor auf allen aktuellen Rechnern interpretiert und angezeigt werden. Zwar bietet eine solche digitale Form durch die erwähnte Suchfunktionalität schon Vorteile gegenüber den gedruckten Büchern, ihrewirklichen Mehrwerte können digitale Wörterbücher aber erst entfalten, wenn die Orientierung ander gedruckten Seite verlassen wird. Als Folge davon wurden in den letzten Jahrzehnten Wörterbuchkonzepte neu bewertet und weiterentwickelt. Die Fortschritte auf dem Gebiet der digitalen Speicherungsmöglichkeiten führten dabei zu immer umfangreicheren Änderungen der Sicht auf lexikalische Einträge.

## 2. Technische Aspekte bei der Speicherung lexikalischer Daten

Digitale Wörterbücher sindnicht mehr nur an der Papierform orientiert, sondern können auch als eigenständige Informationseinheiten betrachtet werden. Wenn Wörterbücher als abstraktere textuelle Objekte behandelt werden, finden dieselben Herangehensweisen an die Speicherung komplexerer Informationseinheiten Anwendung,die auch für andere Informationsobjekte verwendet werden. Es ist z.B. möglich, Wörterbücher als reine Textdateianzusehen und die Speicherung im Dateisystem des Computers vorzunehmen. Bei einem solchen, eher naiven

---

Vorgehen würden eine Reihen von Strukturierungsmöglichkeiten, die die moderne Datentechnik bereitstellt, nicht ausgenutzt.

Grundsätzlich sollten Informationseinheiten bei der elektronischen Speicherung in solchen Systemen repräsentiert werden, die die Daten nicht nur aufbewahren, sondern sie auch durchsuchbar, manipulierbarund weiter verarbeitbar werden lassen. Es ist z. B. erstrebenswert, die Informationseinheiten so zu strukturieren, dass der Verarbeitungsprozess ‚Druck' für das Formatierendes gedruckten Wörterbuchs auf Strukturierungselementen aufsetzen kann, die in die Wörterbuchdaten integriert wurden.

Wörterbücher werden heute in elektronischen Systemen gespeichert und dazu werden – mit Datenbanksystemen –meist Technologiengenutzt, die immer dann genutzt werden, wenn große Datenmengen gespeichert und verarbeitet werden müssen. Datenbanken bieten die Möglichkeit, große Datenmengen in einer strukturierten Form zu speichern. Eine laienhafte Sichtweise auf Datenbanken geht häufig davon aus, dass eine Datenbank eine große Halde darstellt, in der durch quasi magische Prozesse Informationseinheiten sehr schnell herausgelesen und mit anderen Informationseinheiten in Beziehung gesetzt werden können. Natürlich ist die praktische Umsetzung weitaus komplexer. Bevor eine Datenbank aufgesetzt werden kann, müssen die zu speichernden Informationsobjekte und deren Beziehungen zueinander sehr genau analysiert und dann eine Lösung für deren Repräsentation erarbeitet werden. Diese Aufgabe wird von informatisch gebildeten Datenbankexpertinnen und -experten übernommen. Meist kann allerdings nicht davon ausgegangen werden, dass diese Personen auch ein umfangreiches lexikographisches Wissen mitbringen. Um von der Struktur eines Lexikons zur Struktur einer Datenbank zu gelangen, müssen deshalb methodologische Wege gefunden werden, wie die lexikographischen Daten in mehreren Schritten strukturiert werden, so dass sie in einem Datenbanksystem gehalten und verarbeitet werden können. Grundbedingung für eine solche Herangehensweise ist es, dass das lexikographische Wissen in einer formalen Repräsentation abgespeichert werden kann. Im Folgenden wird dargestellt, wie die Speicherung in elektronischen Systemen vorbereitet und lexikographischumgesetzt werden kann.

Ziel bei der Methodenentwicklung für die Speicherung lexikographischer Daten muss es sein, einen lexikographischen Ansatz zu wählen und diesen schrittweise so zu formalisieren, dass alle Informationen, die sich direkt oder indirekt in einem gedruckten Wörterbuch finden lassen, auch repräsentierbarsind. Hierfür werden zwischen der eigentlichen Speicherung und dem zu repräsentierendem Wörterbuch Schichten modelliert, die es erlauben, die Wörterbuchinformationen nicht nur in einer formalen Repräsentationzu speichern, sondern auch die gewünschten Mehrwerte erreichbar werden zu lassen. Wie diese entsprechenden In-formationsobjekte letztendlichdann tatsächlich in einer Datenbank gespeichert werden, ist jedoch nicht Gegenstand dieses Beitrages. Grundsätzlich ist es möglich, die formalisierten Einheiten eins-zu-eins in einer Datenbank zu speichern oder durch Zwischenschritte auf entsprechende Datenbankformate zu übertragen. Dies ist meist nicht die Aufgabe der Lexikographen bzw. der Lexikographinnen, da sie auf semiformalisierten Repräsentationen aufsetzen können sollten.

Bei der Speicherung von Wörterbuchdaten finden heute standardisierte Formate in XML eine sehr weite Verbreitung. Diese Speicherungsform steht deshalb im Zentrum dieses Beitrags. Es ist jedoch auch möglich, lexikographische Daten im Rahmen des sogenannten Ressource Description Framework darzustellen und diese dann in einer Datenbank zugänglich zu machen. Diese Speicherungsmöglichkeit, die besonders für die Repräsentation formaler Ontologien derzeit immer häufiger verwendet wird, soll jedoch ebenfalls nicht weiter in diesem Beitrag thematisiert werden.

Die XML-Repräsentationen erlauben es, in einer idealtypischen Weise vom lexikographischen Wissen hin zu dessen Speicherung zu gelangen. XML ist eine am Text orientierte Auszeichnungssprache, die es erlaubt, Annotationen, die insbesondere zur Beschreibung struktureller Informationen genutzt werden, in einem Textformat aufzunehmen und formal mit Hilfe von Parsern zu überprüfen. Spezielle XML-Editoren erlauben es zudem, den Erstellungsprozess zu unterstützen und die Überprüfung der formalen Struktur schon während der Eingabe des Lexikoneintrags durchzuführen. Mit Hilfe dieser XML-Strukturen können dann die Lexikoneinträge gespeichert und in andere Formate überführt werden. Die eigentlichen Aufgaben der Datenbank, d.h. insbesondere der schnelle Zugriff auf lexikalische Einheiten, ihre Verarbeitung und das Zueinander-in-Beziehung-Setzen von ihnen, kann dann mit Hilfe von regulären Datenbanktechnologien erreicht werden. Aus der Sicht der Lexikographen bzw. Lexikographinnen reicht es allerdings aus zu wissen, dass nachgelagerte Prozesse diese eigentlichen Anwendungen dann umsetzen können.

## 3. Modelle für lexikalische Daten

### 3.1 Zwei komplementäre Modelle

Die Darstellung lexikalischer Daten, unabhängig davon, ob sie in gedruckten Büchern oder in digitalen Ausgaben erscheinen, basiert immer auf der Anwendung eines Modells. Ein derartiges Modell bestimmt die Art und Weise, in der die lexikalischen Einträge repräsentiert werden, z. B. wie die Beziehungen eines Lexems zu seinen Bedeutungen oder zu anderen Lexe-men dargestellt werden oder ganz allgemein wie feingliedrig die entsprechenden Einträge repräsentiert werden sollen.

In diesem Abschnitt betrachten wir zwei unterschiedliche Klassen von Modellen, die in einem komplementären Verhältnis zueinander stehen. Diese Modelle können in gewisser Weise als Metamodelle bezeichnet werden, mit deren Hilfe sich unterschiedliche Arten von Lexikonrepräsentationen bzw. unterschiedliche theoretische Herangehensweisen an die Repräsentation von Lexika abbilden lassen. Die beiden nun betrachteten Modelle bzw. Metamodelle sind die onomasiologischen und semasiologischen Modellklassen.

Die semasiologischen Modelle beziehen sich auf eine Darstellung, in der das Lemma, gegeben als Sequenz von Buchstaben, Ausgangspunkt für die Repräsentation ist. Jedem Lemma wird eine Bedeutung oder ggf. mehrere Bedeutungenzugeordnet.

Im Gegensatz dazu steht das onomasiologische Modell, bei dem die Bedeutungen bzw. die Konzepte der entsprechenden lexikalischen Einheiten im Zentrum der jeweiligen Repräsentationenstehen. Bei den onomasiologischen Modellen ist es daher direkt möglich, ein Konzept in unterschiedlichen Sprachen darzustellen, d. h. einem Konzept können nicht nur alle dazugehörigen synonymen Wörter in einer Sprache zugeordnet werden, sondern es können von demselben Konzept auch Verweise auf die entsprechenden Lexikoneinträge oder Wörterbucheinträge in anderen Sprachen vorhanden sein. Die onomasiologischen Modelle werden für Thesauri, Synonymwörterbücherund für terminologische Wörterbücher verwendet, bei denen jeder Eintrag genau einer Bedeutung entspricht. Daher finden die auf Basis der onomasiologischen Modelle erstellten Wörterbücher besonders im Übersetzungswesen häufig ihre Verwendung.

Trotz ihres grundsätzlich unterschiedlichen Ansatzes und ihrer gegensätzlichen Herangehensweise in Bezug auf die Anordnung der Daten können sowohl auf Grundlage onomasiologischer als auch auf Basis semasiologischer Modelle lexikalische Relationenwie Polysemie und Synonymie dargestellt werden. In den folgenden Abschnitten werdenan Hand dieser Relationenbeide Modelle etwas genauer dargestellt. Später werden wir zeigen, wie diese Modelle als Basis für die maschinelle Repräsentation und die maschinelle Speicherung von Lexikoneinträgen verwendet werden können.

## 3.2 Polysemie

Bei polysemen Termen werden unterschiedliche Bedeutungen mit ein und demselben Lexikoneintrag in Verbindung gebracht. Die semasiologischen Modelle sind so aufgebaut, dass bei polysemen Wörtern die mehrfachen Bedeutungen, die einem Lemma zugeordneten wer-den können, direkt mit dem Lexikoneintrag desentsprechendenLemmasverbunden sind. Daraus folgt, dass jedem Lexikoneintrag 1 bis n verschiedene Bedeutungen zugeordnet werden können. Die meisten gedruckten Wörterbücher und Enzyklopädien sind gemäß diesem Organisationsprinzip aufgebaut. Aber auch bei elektronisch gespeicherten ein- und mehrsprachigen Wörterbüchern ist diese Struktur häufigzu finden. Wie ausführlich die Bedeutungen eines lexikalischen Eintrags beschrieben werden oder wie fein die Granularität bzw. wie umfangreich die Auswahl der entsprechenden Bedeutungen gestaltet wird, ist natürlich eine redaktionelleEntscheidung, die von Wörterbuch zu Wörterbuch unterschiedlich ist.

DieBehandlung von Polysemie ist eine größte Herausforderung für die onomasiologischen Modelle, da sieerst einmal keineMöglichkeiten besitzen, diese direkt abzubilden. Der Grund hierfür liegt darin, dass die Bedeutung eines bestimmten Terms die Basis einer Repräsentation ist. Wenn es jedoch mehrere Bedeutungen gibt, ist dies in der Repräsentationsstruktur nicht sofort ersichtlich. Diese Problematik soll durch ein Standardbeispiel veranschaulicht werden: Das Wort „Absatz" hat im Deutschenmehrere Bedeutungen: Im Buchdruckwesensteht es für einen Paragraph oder einen Abschnitt; in der eher wirtschaftlich ökonomisch buchhalterischen Herangehensweise wird es als Produktabsatz gesehen; im Kontext der Beschreibung chemischen Prozessen kann das Wort „Absatz"Ablagerungen oder Rückstände beschreiben; in alltäglicher Umgangssprache benennt ein „Absatz" einen Teil der Schuhsohle. Diese verschiedenen – aberdurch dasselbe polyseme Wortbeschriebenen – Konzeptewerden in der onomasiologischen Repräsentation vollständig unabhängig voneinander dargestellt. Nur durch Mechanismen wie Querverweise können die Verbindungen zwischen polysemen Ausdrücken dargestelltwerden.

## 3.3   Synonymie

Bei der Synonymie verhält es sich für die Speicherung lexikalischer Terme in gewisser Weise exakt komplementär zu der Speicherung im Falle der Polysemie. Die Modellierung von Synonymie ist eineAufgabe, die mit einer onomasiologischen Herangehensweise einfacher angegangen werden kann. Bei den onomasiologischen Modellen steht die Bedeutung im Vordergrund und die verschiedenen Wörter, die verwendet werden können, um dieses Konzept auszudrücken, werden dann dem entsprechenden Konzeptzugeordnet. Somit werden praktisch den einzelnen Konzepten in Abhängigkeit von der Modellierungstiefe und dem Umfang des entsprechenden Wörterbuches sämtliche synonymeAusdrücke,d.h. Wörter, zugeordnet, die dieses Konzept bezeichnen. Die Synonyme eines Wortes $w_1$ sind somit dann alleWörter $w_2$ – $w_n$, diedemselben Konzept zugeordnet sind. Auf diese Weise kann die gesamte Menge von

Wörtern nach Sachgruppen und Bedeutungsverwandtschaften gegliedert werden. Die meisten klassischen Synonymwörterbücher, aber auch Wortnetze wie z.B. WordNet[9] oder Germa-Net[10], werden gemäß dieser Herangehensweise repräsentiert.

Bei dem onomasiologischen Modell können zudem noch weitere Informationen hinzugefügt werden, z.B. können die präferierten Terme hervorgehoben werden oder es werden entsprechende fachsprachliche Ausdrücke oder bestimmte Register bei der Auswahl der entsprechenden Terme im Wörterbuch vermerkt. Wie im Folgenden beschrieben wird, ist der onomasiologische Ansatz auch für mehrsprachige Wörterbücher geeignet. Zudem wird er auch bei terminologischen Datenbanken, einem wichtigen Hilfsmittelsowohl für menschlicheÜbersetzer und als auch für Übersetzungssoftware, verwendet.

## 3.4    Aspekte der Multilingualität

Die Repräsentation von lexikalischen Einheiten, diefür multilinguale Wörterbücher vorgesehen sind, kann durch semasiologischeoderonomasiologischeModelle vorgenommen werden. Dabei ähnelt die Modellierung derRepräsentationvon Synonymen. Bei dersemasiologischen-Modellierung muss man für die verschiedenen sprachlichen Varianten der Übersetzungen entsprechende Bedeutungen angeben. Das führt dazu, dass für jedes Wort der einen Sprache gemäß ihrer Bedeutungenmehrere unterschiedliche Wörter der anderen Sprache zugeordnet werden. Besonders problematisch wird dieser Ansatz, wenn man nicht nur zweisprachige, sondern auch multilinguale Wörterbücher mit drei oder mehr Sprachen betrachtet. DieseProblematik ergibt sich aus der kombinatorischen Explosion.

Im Gegensatz zu den semasiologischen Modellierungen eignen sich onomasiologische Modellierungen idealtypisch für die Repräsentation von multilingualen Lexika und Wörterbüchern. Für ein gegebenes(lexikalisches) Konzept erlaubt dieses Modell die verschiedenen Einträge gemäß ihrer Sprache zu repräsentieren. Dies kann bei einer passenden Konzeptwahl so gestaltet werden, dass beliebig viele Sprachen modelliert werden können. In der Tat ist es so, dass die Terminologie, die z. B. für die Übersetzung von technischen Dokumentationen verwendet wird, in den entsprechenden Übersetzungsdatenbanken nach dem onomasiologischen Modell repräsentiert wird.

---

[9]    http://wordnet.princeton.edu
[10]    http://www.sfs.uni-tuebingen.de/lsd

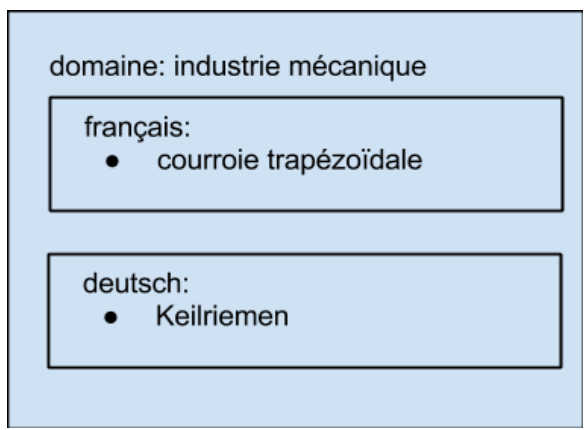

**Abbildung 1:** Datenmodell zur Repräsentation von multilingualen Lexika

## 4. Vom Modell zum Format

Nachdem die Auswahl einer der beiden Herangehensweisen zur Modellierung der lexikalischen Ressourcengetroffen wurde, muss entschieden werden, welche Repräsentation zum Speichern der Daten verwendet werden soll, um dann die lexikalischen Daten entsprechend abbilden zu können. Hierfür ist es notwendig, ein formalisiertes Modell zu entwickeln, das die Informationen, die im theoretischen Modell repräsentiert werden, vollständig ausdrücken kann. Nachdem diese grundlegenden Modellierungsarbeiten geleistet wurden, ist es vergleichsweise einfach, diese Modelle in eine formale Sprache zu überführen.

Grundsätzlich kann man sowohl XML-basierte als auch nicht XML-basierte Modellierungssprachen verwenden, um verschiedene Modellierungsansätzen zurealisieren. Dabei sollte jedoch immer ein mathematischesModelldie Grundlage für die Modellierung bilden. Die in diesem Kontext wichtigsten Modelle sindListen, Bäumesowie allgemeine Graphen. Letztere erlauben es, mit sogenannten Kanten von einem Knoten, der eingespeichertesInformationsobjekt repräsentiert, auf einen oder mehrere andere Knoten zu verweisen. Fragen, die sich bei diesem Prozess zu stellen sind, sind z.B.: Stellt eine Baumstruktur eine adäquate Modellierung der Daten dar? Kann ich mein Wissen in Form von flachen Listen ausdrücken?Benötige ich allgemeine Graphen?

Wenn ein Modell evaluiert und als zur Abbildung der darzustellenden linguistischen Phänomene geeignet eingestuft wurde,wird eine Modellierungssprache gewählt (bzw. neu definiert), die es erlaubt, die entsprechenden Modellierungsgrundsätze auszudrücken. Fällt die Wahl auf eine XML-basierte Modellierungssprache, werden die Modellierungsdetails mittels einer Dokumentgrammatik festgelegt. DieDokumentgrammatik wird z.B. verwendet, um die Lexikoneinträge in einer adäquaten Form zu repräsentieren und den Editierungsprozess ma-schinell zu steuern. Letzteres vereinfacht nicht nur die Arbeit der Redakteurinnen und Redakteure,sondern trägt auch dazu bei, Strukturierungsfehler bei der Erstellung der Wörterbücher auszuschließen. Das Ausschließen von formalen Fehlern erfolgt mit Hilfe von Parsern, die die entsprechenden Lexikoneinträge, die dann als sogenannteXML-Instanzen repräsentiert sind, gegen die entsprechenden XML-Grammatiken validieren und feststellen, ob es formale Fehler in der Repräsentation dieser Instanzen gibt. Wenn dieser Prozess erfolgreich war, dann liegt zusätzlich die elektronische Repräsentation eines gesamten Lexikons in Arbeitsspeicher. Der Prozess des Herauslesens erlaubt dann die gewünschtenLinearisierungen bzw. Filterungen der Daten vorzunehmen, um z. B. das Wörterbuch oder den Lexikoneintrag zu generieren.

Ein weit verbreitetes Repräsentationsmodell zur Speicherung von elektronischen Daten in der Linguistik, aber auch in anderen Geisteswissenschaften, bietet das Modell der TEI. Die TEI hat sich im Verlauf ihrer mittlerweile mehr als zwanzigjährigen Geschichte auch mit der Problematik der Repräsentation von Wörterbüchern und von lexikalischen sowie terminologischen Daten beschäftigt.

Die klassische Repräsentationsform für Wörterbücher gemäß der Herangehensweise der TEI orientierte sich an gedruckten Wörterbüchern, die dann im Zuge einer Retro-Digitalisierungin elektronische Ressourcen überführt werden sollten. Heute bietet die TEI allerdings auch Möglichkeiten, sie für lexikalische Repräsentationen zu verwenden, die schon in elektronischer Weise aufgebaut werden, also solche, die keinen Vorläufer in einer gedruckten Form besitzen.

Für terminologische Datenbanken hatte die TEI in ihren Fassungen P3 und P4 auch einen Mechanismus bereitgestellt, der aber in den zur Familie P5gehörenden aktuellen Fassungen nicht mehr vorhanden ist. Für diese Daten werden heute Standards der Internationalen Standardisierungsorganisation ISO verwendet, die im nächsten Abschnitt vorgestellt werden.

Die Abbildung 2 stellt die Beziehung der Modelle zur Repräsentation von Wörterbuchdaten schematisch dar. Sowohl für semasiologische als auch für onomasiologische Herangehensweisen existieren abstrakte Modelle und konkrete Linearisierungen, die jeweils in den nächsten Abschnitten kurz dargestellt werden. Eine weitere Komponente der Modellierungsmöglichkeiten, auf dein diesem Beitrag jedoch nicht eingegangen wird, bilden die Datenkategorien, die von sehr vielen für Sprachdaten relevanten Standards genutzt werden.

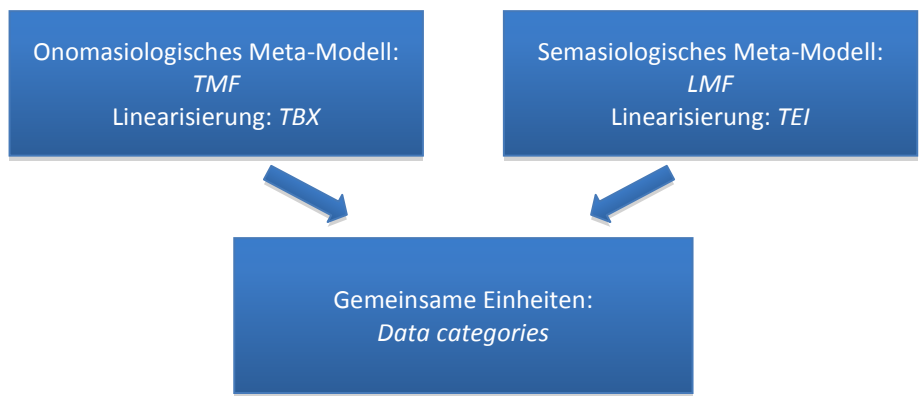

**Abbildung 2:** Modellierungsansätze für Wörterbuchdaten

## 5. Standards und Formate für onomasiologische Modelle

### 5.1 Grundlegende Prinzipien

Der Repräsentation terminologischer Einträge gemäß demonomasiologischen Modell liegen zwei Basisprinzipien zugrunde:

– Orientierung an der Konzeptebene (concept orientation):Alle Informationen, die zu einem Konzept gehören,werden diesem Konzept beigeordnet. Im Falle mehrsprachiger Wörterbücher werden in ein Konzeptalle Sprachen aufgenommen, d. h. die verschiedenen Wörter der unterschiedlichen Sprachen werden dem einen Konzept zugeordnet.

– Term-Autonomie (term autonomy): Dieses Prinzip besagt, dass jeder einzelne Term aus sich selbst heraus erklärbar sein muss, was de facto dazu führt, dass alle Terminologie-Einträge, die in einem derartigen System betrachtet werden, für sich alleine stehen können.

### 5.2 Das Meta-ModellTMF

Der ISO-Standard 16642 (Terminological Markup Framework, TMF) definiert ein Meta-Modell, das unterschiedliche Modelle, die lexikalischen Einträge in einer onomasiologischen Herangehensweise repräsentieren können,beschreibt. Es wird insbesondere für mehrsprachige Terminologien verwendet. Zur Illustration sei auf die Abbildung 3zu dem Lexikoneintrag ‚Keilriemen‘ verwiesen. Dieses Konzept wird in dem terminologischen Standard ISO 16642 in einer onomasiologischen Weise abgebildet. Dabei werden die Informationen auf drei verschiedenen Ebenen dargestellt: auf der konzeptuellen Ebene (terminological entry), auf der Sprachebene und auf der Termebene. Die konzeptuelle Ebene umfasst eine Definition des entsprechenden Terms sowie gegebenenfalls Relationen dieses Terms zu anderen Konzepten, z. B. Synonymie oder Meronymie. Die Ebene der Sprache (language section), gruppiert die einzelnen Übersetzungen oder einzelnen Bezeichnungen in den verschiedenen Sprachen und gegebenenfalls auch in regionalen Varianten dieser Sprachen und legt sie in dem Terminologiesystem ab. Die Termebene gruppiert all die Informationen, die die Terme beschreiben und sorgt dafür, dass das Prinzip der Autonomie der terminologischen Einträge eingehalten wird.

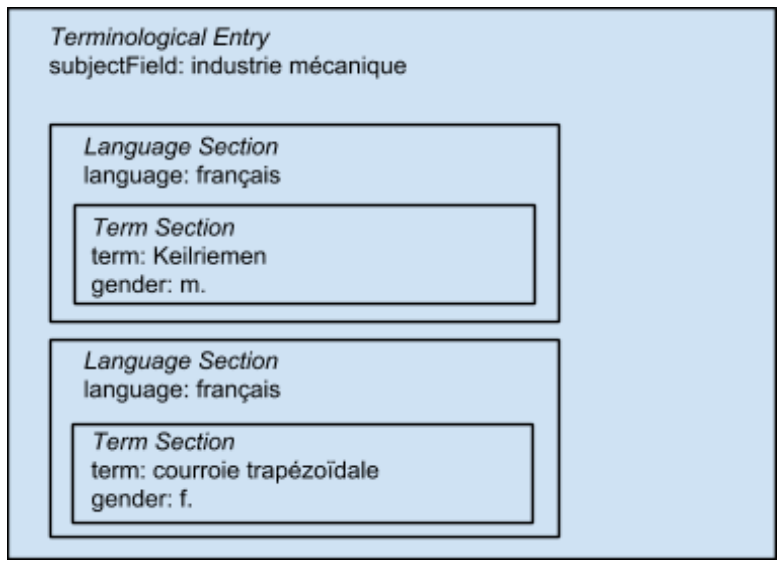

**Abbildung 3:** TMF-Metamodell für einen terminologischen Eintrag

Die überwiegende Mehrheit der Kategorien für diese terminologischen Datenbanken ist bereits in dem älteren Standard ISO 12620 aus dem Jahre 1999 beschrieben worden. Der Standard ISO 16642 nutzt die in ISO 12620 definierten Konzepte.

Um ein gültiges terminologisches Modell zu erhalten, ist es notwendig, zumindest zwei Kategorien anzugeben. Diese obligatorischen Kategorien, die den Term beschreiben, sind (1) die Referenz auf das Lemma und (2) die Sprachsektion, in der die entsprechende sprachliche Repräsentation des entsprechenden Terms wiedergegeben wird (s. Abb. 4). Typischerweise finden sich dann verschiedene Sprachsektionen zu den einzelnen Termen, wodurch die Repräsentation multilingualer Lexika möglich wird. Um einen terminologischen Eintrag möglichst gut und elaboriert zu beschreiben, stehen verschiedene Kategorien zur Verfügung. Diese stammen aus der Norm ISO 12620. Erstens gibt es eine Kategorie ‚subjectField‘, dieeineBeschreibung der Domäne, also z. B. eines Fachgebietes bei fachterminologischen oder fachsprachlichen Einträgen erlaubt. Zweitens gibt es die Kategorie ‚Definition‘, in der für Einträge eine entsprechende Definition beigefügt wird. Meist ist das eine Definition, die gemäß der ISO-Norm 704 (Terminologiearbeit – Grundsätze und Methoden) aufgebaut ist. Und schließlich gibt es drittens die Kategorie ‚partOfSpeech‘, möglicherweise ergänzt um eine entsprechende Genus-Angabe, in der die entsprechenden Wortartenangegeben werden.

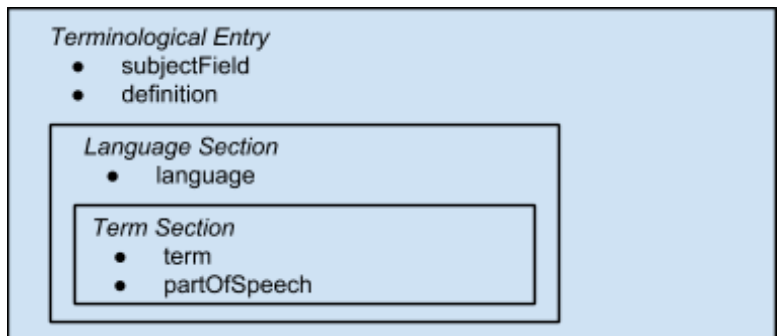

**Abbildung 4:** TMF-Metamodell mit Sprach- und Termbeschreibung

## 5.3   Serialisierung des Formates TMF mit Hilfe des Formates TBX

Der Standard ISO 16642 (TMF) ist das Ergebnis einer sehr langen Entwicklung in der Normungsarbeit. Seit den 1980er Jahren sind hierzu eine Vielzahl von Publikationen erschienen. TMF beschreibt die konzeptuelle Ebene. Allerdings ist es ebenfalls notwendig, die entsprechenden terminologischen Einträge auch in einer serialisierten bzw. linearisierten Form auszudrücken;nicht zuletzt, um sie in verschiedenen technischen Systemen zu nutzen, d.h. u.a. sie exportieren und importieren, zu können. Für diesen Zweck ist die Norm ISO 30042 (TBX) entwickelt worden. In gewisser Weise definiert TBX eine XML-Fassung für terminologischen Einträge, die TMF-konform repräsentiert wurden. Basis des Metamodells TMF ist der terminologische Eintrag `<termEntry>`sowie die Sprach- und Termsektion. Diese Einheiten werden in TBX mit Hilfe von XML kodiert und auf die XML-Elemente `<termEntry>`,`<longSet>`und `<tig>` abgebildet. Nachfolgend sei eine sehr einfache in gewisser Weise strukturell minimale Repräsentation eines Terms angegeben; in diesem Fall der Term ‚e-mail‘:

```
<termEntry xml:id=”c5” xmlns=”http://www.tbx.org”>
<langSet xml:lang=”en”>
<tig>
<term>e-mail</term>
</tig>
</langSet>
<langSet xml:lang=”fr”>
<tig>
<term>courriel</term>
</tig>
</langSet>
</termEntry>
```

Die anderen Kategorien werden mit Hilfe von generellen Elementen oderAttributen, wie z.B. dem Attribut `@type`, repräsentiert. Das Element`<descrip>`wird verwendet, um Informationen,die einen Definitionscharakter haben,für einen entsprechenden terminologischen Eintragzusammenzufassen. Mithilfedes Elements `<admin>`werdenadministrative Informationen erfasst und das Element`<termNote>`wird verwendet, umweitere, spezifischere Informationen einem Term zuzuordnen. Im folgendenBeispiel kann man die verschiedenen Elemente in ihrer Anwendung sehen:

```xml
<termEntry>
<descripGrp>
    <descrip type="originatingDatabaseName">Vocabulaire multidisciplinaire PAS
    CAL</descrip>
<descrip type="subjectField" xml:lang="fr">Biomédical</descrip>
<admin type="conceptIdentifier">BV.122497</admin>
<admin type="conceptOrigin">INIST</admin>
</descripGrp>
<langSet xml:lang="fr">
<tig>
<term>Acide 1,2-dithiolane-3-valérique</tei:term>
<admin type="termIdentifier">BV.122497.1</admin>
<termNote type="administrativeStatus">preferredTerm</termNote>
</tig>
<tig>
<term>«1,2»-Dithiolane-«3»-valérique acide</tei:term>
<admin type="termIdentifier">BV.122497.2</admin>
<termNote type="administrativeStatus">deprecatedTerm</termNote>
</tig>
<tig>
<term>Acide α-lipoïque</tei:term>
<admin type="termIdentifier">BV.122497.3</admin>
</tig>
<tig>
<term>Acide thioctique</tei:term>
<admin type="termIdentifier">BV.122497.4</admin>
</tig>
</langSet>
<langSet xml:lang="en">
<tig>
<term>Thioctic acid</tei:term>
<admin type="termIdentifier">BV.122497.5</admin>
<termNote type="administrativeStatus">preferredTerm</termNote>
</tig>
</langSet>
</termEntry>
```

## 6. Standards und Formate für semasiologische Modelle

### 6.1 Grundlegende Prinzipien

Die relevanten Modellierungsmöglichkeiten für die digitale Repräsentation von Lexika sind derzeit Formate, die in einer modularen Weise aufgebaut sind. Hierbei sind natürlich als Erstes die Arbeiten der Text Encoding Initiative (TEI) zu nennen. Als weiterer Standard, der insbesondere für die Repräsentation gedruckter Wörterbücher entwickelt wurde, existiert der ISO-Standard 1951, der allerdings hier nicht näher betrachtet werden soll.

Ebenfalls von der internationalen Standardisierungsorganisation ISO wurde mit dem Lexical Markup Framework (LMF, ISO 24613) ein weiterer Formalismus entwickelt. Nachfolgend werden LMF und das TEI-Modul für Wörterbücher kurz beschrieben (s. auch Lemnitzer et al. 2013).

## 6.2 Der ISO-Standard LMF

Der Standard LMF bildet einen Metastandard für die Repräsentation lexikalischer Daten. Die LMF-Kernkomponenten beschreiben die elementare Hierarchie lexikalischer Informationen. Zusätzlich hierzu definiert LMF u. a. Erweiterungen für die Repräsentation maschinell lesbarer Lexika.

Da LMF ein Metamodell für die lexikalischen Einträge definiert, bietet es Möglichkeiten der abstrakten Repräsentation lexikalischer Informationen. Das Hauptanwendungsfeld von LMF sind Wörterbücher, die für die maschinelle Verarbeitung natürlicher Sprache verwendet werden. Neben der Beschreibung ihrer abstrakten Modellierungsherangehensweise bietet die Norm LMF auch eine exemplarische Linearisierung. LMF selbst ist allerdings in der Modellierungssprache Unified Modeling Language (UML) beschrieben, die in der Informatik entwickelt wurde und meist auch in der Informatik verwendet wird.UML erlaubt es, einzelne Informationseinheiten zu repräsentieren, und bietet auch Möglichkeiten, diese Repräsentation in XML zu linearisieren. Da eine mögliche XML-Linearisierung exemplarisch im Anhang des LMF-Standards beschrieben ist, wird dieseBeschreibung häufig als die LMF-konforme Repräsentation für lexikalische Einheiten gewählt. Dabei wird übersehen, dass auch andere XML-basierte oder auch nicht XML-basierte Repräsentationen durchaus LMF-konform sein können.

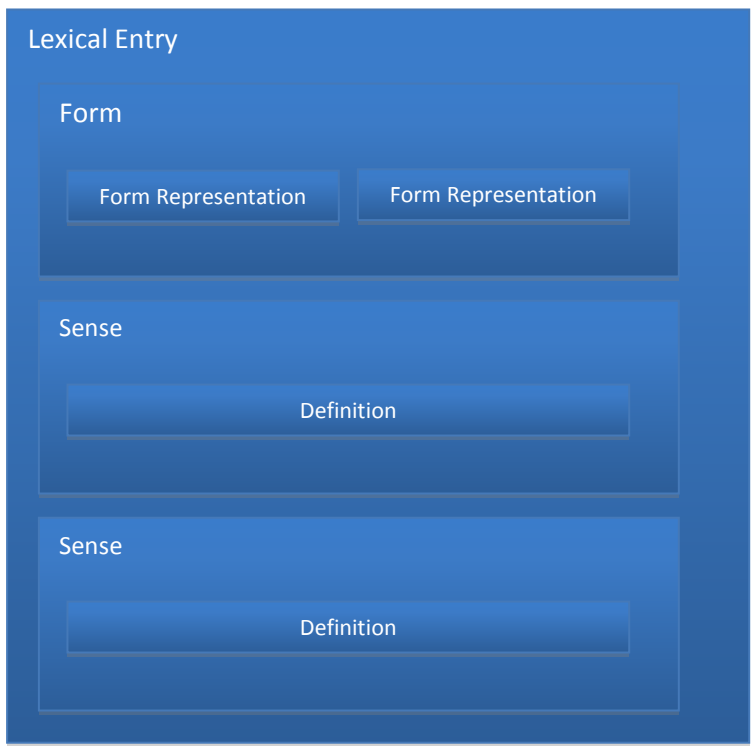

**Abbildung 5:** LMF-Metamodell für einen lexikalischen Eintrag

## 6.3 TEI

Die Text Encoding Initiative bildet die vermutlich wichtigste Standardisierungseinheit für die Repräsentation geisteswissenschaftlicher Informationen. Die TEIhat eine Vielzahl von Modulen für verschiedene Dokumentarten definiert (vgl. Burnard/Bauman 2014). Anfang der 1990er Jahre wurden die ersten XML basierten Schemata zur Repräsentation von geisteswissenschaftlichen Texten durch die TEI veröffentlicht. Zu den Dokumenttypen, die durch die TEI unterstützt werden, gehören u. a. Prosatexte, Lyrik, Texte der Gattung Drama, Manuskripte und eben auch Wörterbücher. Die TEI hat sich im Laufe der Jahre mehr und mehr zu einer Infrastruktur entwickelt, die es erlaubt, bestimmte Komponenten zu definieren, hinzuzufügen oder auszublenden.

    In dem TEI-Modul zur Repräsentation von lexikalischen Einheiten, die sich an den gedruckten Wörterbüchern orientieren, stehen insbesondere zwei Elemente zur Verfügung, die für die Kodierung von Lexikoneinträgen verwendet werden können. Zum einen gibt es das Element `<entryFree>`, das es erlaubt, in einer selbst definierten oder zumindest relativ freien Art und Weise lexikalische Einheiten oder Lemmata wiederzugeben, so wie sie sich auch in einem Wörterbuch befinden. Dieses Element kann beispielsweise dann sehr hilfreich sein, wenn z.B. atypisch strukturierte Wörterbücher, die in vergangenen Jahrhundertenerschienen

sind, in einer XML-kompatiblen Form repräsentiert werden sollen. Das andere Element, das die TEI zur Verfügung stellt, ist das Element `<entry>`, das es erlaubt,in einer stark strukturierten Form – vielleichtam ehesten vergleichbar mit den Elementen, die sich in einer Datenbank befinden – Lexikoneinträgezu repräsentieren.

Die Unterscheidung von zwei verschiedenen Elementen bei der Repräsentation der Lexika bietet zwei Vorteile: Einerseits gibt das Element `<entryFree>` eine relative Freiheit, bei der atypisch strukturierte Lexika oder Wörterbücher repräsentiert werden können; andererseitsbietet das Element `<entry>`mit seiner vergleichsweise starren Herangehensweise eine Hilfestellung, die es ermöglicht, sehr detaillierte und hoch strukturierte Einheiten zu repräsentieren.

Der TEI-Ansatz zur Repräsentation von Wörterbuchdaten ist in gewisser Weise ein prototypischer Vertreter für die semasiologische Sicht auf lexikalische Einträge und kann sehr gut auf ein LMF konformes Repräsentationsmodell abgebildet werden. Dies wird dadurch ermöglicht, dass die TEI insbesondere zwei Konstruktionen bzw. zweiweitere Elemente erlaubt, die in dem Entry-Element vorkommen.

Das erste Element ist `<form>`, das weitere deskriptive Elemente umschließen kann. Esermöglicht die Angabe von Informationen zur Aussprache, Orthographie, Grammatik etc. in einer oberflächennahen Repräsentation. Auf diese Weise kann innerhalb des Elementes `<form>` durch das Element `<orth>` dargestellt werden, dass es unterschiedliche orthographische Varianten für ein Wort gibt. Durch das Element `<gram>`können verschiedene grammatische Informationen wie Wortart, Geschlecht etc. wiedergegeben werden.

Das zweite Element `<sense>` erlaubt, die potentiell verschiedenen Bedeutungen, die mit einem Wort, mit einem Eintrag verbunden sind, wiederum inhierarchischer Weise zu repräsentieren. Darüber hinaus ermöglicht die Repräsentation der verschiedenen Bedeutungen innerhalb von `<sense>`auch, eine genauere Spezifikation der Form vorzunehmen. So ist es möglich, dass bestimmte Formen nur mit bestimmten Bedeutungen verbunden sind. Manche Bedeutungen können z. B. nur in der Singularform eines Eintrags repräsentiert werden. Innerhalb des Sense-Elementes können Informationen wiedergegeben werden, die die Bedeutungen eines Worts betreffen, z.B. in dem Element `<def>`eine Definitionder Bedeutung.Mit dem Element `<cit>` können Belege für die sprachliche Verwendung des Wortes gegeben werden. Verwendungen, die eher einen informellen Charakter haben, können mit dem Element `<exemplum>` angegeben werden.

Durch einen sogenannten „usage marker", der durch das Element `<usg>` kodiert wird, kann man einem Wort weitere Informationenzuordnen. Diese Beschreibung kannz.B. Infor-mationen zu bestimmten regionalen Varianten oder der entsprechenden diachronen Beschreibung eines Wortes enthalten. So kannz. B. so etwas wie ‚veraltet' modelliert werden, d.h. das Wort findet in der zeitgenössischen Ausprägung der Sprache keine Verwendung mehr.

Das folgende Beispiel veranschaulicht eine vollständige, LMF-konforme TEI-Repräsentation des Wortes „poussin" aus einem französischen, einsprachigen Wörterbuch. Unter einemLexikoneintrag werden mehrere Bedeutungen angeführt, die,je nachdem im welchem Kontext die lexikalische Einheit „poussin"verwendet wird, variieren:

```xml
<entry>
<form>
<orth>poussin</orth>
<pron>pusë</pron>
</form>
<gramGrp>
<pos>n.</pos>
<gen>m.</gen>
</gramGrp>
<sense n="1">
<def>Jeune poulet, nouvellement sorti de l'oeuf, encore couvert de duvet.</def>
<cit type="example">
<quote>La poule et ses poussins.</quote>
</cit>
</sense>
<sense n="2">
<usg type="dom">Zool.</usg>
<def>Jeune oiseau (par rapport aux adultes, aux parents)</def>
</sense>
<sense n="3">
<usg type="dom">êtres humains</usg>
<sense n="3.1">
<usg type="register">Fam.</usg>
<def>Terme d'affection (enfant)</def>
</sense>
<sense n="3.2">
<usg type="dom">Sports</usg>
<def n="2">Catégorie d'âge (9 ans) qui précède celle des benjamins</def>
</sense>
<sense>
<def n="3">Elève de première année dans certaines écoles (Air, Aéronautique)</def>
</sense>
</sense>
</entry>
```

## 7. Zusammenfassung

Die Methoden zur Speicherung lexikalischer Daten sind immer abhängig von der Sicht auf Wörterbücher. Nachdem die grundlegenden lexikographischen Entscheidungen getroffen wurden, müssen die lexikalischen Daten genauestens analysiert werden. Erst dann kann eine Form der Datenspeicherung gewählt werden. Um die Austauschbarkeit der digitalen Wörterbuchquellen sicherzustellen, sollte die Wahl auf ein standardisiertes Format fallen, das ggf. nach den Projektbedürfnissen modifiziert werden kann.